\documentclass[letterpaper, 10 pt, conference]{ieeeconf}  % Comment this line out if you need a4paper

\IEEEoverridecommandlockouts                              % This command is only needed if 
\usepackage{graphics} % for pdf, bitmapped graphics files
\usepackage{graphicx} 
\usepackage{amsmath} % assumes amsmath package installed
\usepackage{multirow}
\usepackage{amsfonts}
\usepackage [ linecolor = black , backgroundcolor = lightgray ]{ todonotes }

\title{\LARGE \bf
Selecting Mechanical Parameters of a Monopode Jumping System with Reinforcement Learning
}

\author{Andrew Albright$^{1}$ and Joshua Vaughan$^{1}$% <-this % stops a space
\thanks{$^{1}$Andrew Albright and Joshua Vaughan, Mechanical Engineering Department, University of Louisiana at Lafayette, Lafayette, LA 70503
        {\tt\small andrew.albright1@louisiana.edu}}%
}

\begin{document}

\maketitle
\thispagestyle{empty}
\pagestyle{empty}

%%%%%%%%%%%%%%%%%%%%%%%%%%%%%%%%%%%%%%%%%%%%%%%%%%%%%%%%%%%%%%%%%%%%%%%%
\begin{abstract}

Legged systems have many advantages when compared to their wheeled counterparts. For example, they can more easily navigate extreme, uneven terrain. However, there are disadvantages as well, particularly the difficulty seen in modeling the nonlinearities of the system. Research has shown that using flexible components within legged locomotive systems improves performance measures such as efficiency and running velocity. Because of the difficulties encountered in modeling flexible systems, control methods such as reinforcement learning can be used to define control strategies. Furthermore, reinforcement learning can be tasked with learning mechanical parameters of a system to match a control input. It is shown in this work that when deploying reinforcement learning to find design parameters for a pogo-stick jumping system, the designs the agents learn are optimal within the design space provided to the agents.

\end{abstract}

%%%%%%%%%%%%%%%%%%%%%%%%%%%%%%%%%%%%%%%%%%%%%%%%%%%%%%%%%%%%%%%%%%%%%%%%
\section{Introduction}
\label{sec:intro}
The use of flexible components within legged locomotive systems has proved useful for both reducing power consumption and increasing performance \cite{Sugiyama2004, Buondonno2017, Hurst2008}. However, designing controllers for these systems is difficult as the flexibility of the system generates nonlinear models. As such, employing series-elastic-actuators (SEA) instead of flexible links is an attractive and popular solution, since the models of the systems become more manageable \cite{Buondonno2017, Zhang2019, Pratt1995}. Still, the use of SEAs do not represent the full capability of flexible systems. As a result, other methods that use flexible tendon-like materials meant to emulate more organic designs have been proposed \cite{Iida2005}. These, however, are still not representative of fully flexible links, which have been shown to drastically improve locomotive performance measures such as running speed \cite{Saranli2001}.

Control methods have been developed that work well for flexible systems like the ones mentioned \cite{Luo1993, Modeling2003}. However, as the systems increase in dimensionality, effects such as dynamic coupling between members make such methods challenging to implement. As such, work has been done that uses neural networks and methods such as reinforcement learning (RL) to develop controllers for flexible systems \cite{Bhagat2019e, Thuruthelb}. For example, RL has been used to create control strategies for both flexible-legged and rigid locomotive systems that when compared, show the locomotive systems outperform their rigid counterparts \cite{Dwiel2019d}. Furthermore, those controllers were shown to be robust to changes in design parameters. 

In addition to the work done using RL to develop controllers for flexible systems, work has been completed which shows that this technique can be used to concurrently design the mechanical aspects of a system and a controller to match said system \cite{Ha2019j}. These techniques have even been used to define mechanical parameters and control strategies where the resulting controller and hardware were deployed in a sim-to-real process, validating the usability of the technique \cite{Chen2020}. Using this technique for legged-locomotion has also been studied, but thus far has been limited to the case of rigid systems \cite{Schaff2019e}. 

As such, this paper explores of using RL for concurrent design of flexible-legged locomotive systems. A simplified flexible jumping system was used where, for the initial work, the control input was held fixed so that the RL algorithm was tasked with only learning optimized mechanical parameters. The rest of the paper is organized such that in the next section, similar work will be discussed. In Section \ref{sec:pogo_model}, the pogo-stick environment details will be defined. Next, in Section~\ref{sec:control_input_exp}, the input used during training will be explained. Then, in Section~\ref{sec:learning_mech_params}, the algorithm used along with the method of the experiments will be presented. The performance of the learned designs are shown in Section~\ref{sec:results}. % Leaving conclusion section out intentionally

%%%%%%%%%%%%%%%%%%%%%%%%%%%%%%%%%%%%%%%%%%%%%%%%%%%%%%%%%%%%%%%%%%%%%%%%
\section{Related Work}
\label{sec:related_work}
\subsection{Flexible Locomotive Systems}

The use of flexible components within locomotive robotics systems has shown improvements in performance measures such as movement speed and jumping height \cite{Sugiyama2004, Hurst2008}. Previous work has shown that the use of flexible components in the legs of legged locomotion systems increase performance while decreasing power consumption \cite{Saranli2001}. Research also has been done showing the uses of series-elastic-actuators for locomotive systems \cite{Pratt1995}. In much of this work, human interaction with the robotic systems is considered such that rigidity is not ideal \cite{Zhang2019}. The studies of flexible systems are challenging however, as the models which represent them are often nonlinear and therefore difficult to develop control systems for. As such, there is a need for solutions which can be deployed to develop controllers for these nonlinear systems.

\subsection{Controlling Flexile Systems Using RL}

Control methods developed for flexible systems have been shown to be effective for position control and vibration reduction \cite{Luo1993, Ahmadi1997}. Because of the challenges seen in scaling the controllers, methods utilizing reinforcement learning are of interest. This method has been used in simple planar cases, where it was compared to a PD control strategy for vibration suppression and proved to be a higher performing method \cite{He2020f}. Additionally, it has also been shown to be effective at defining control strategies for flexible-legged locomotion. The use of actor-critic algorithms such as Deep Deterministic Policy Gradient \cite{Lillicrap2016h} have been used to train running strategies for a flexible legged quadruped \cite{Dwiel2019d}. Much of the research is based in simulation, however, and often the controllers are not deployed on physical systems, which leads to the question of whether or not these are useful techniques in practice.

\subsection{Concurrent Design}

Defining an optimal controller for a system can be difficult due to challenges such as mechanical and electrical design limits. This is especially true when the system is flexible and the model is nonlinear. A solution to this challenge is to concurrently design a system with the controllers so that the two are jointly optimized. This strategy has been used to develop better performing mechatronics systems \cite{Li2001}.  More recent work has used advanced methods such as evolutionary strategies to define robot design parameters \cite{Wang2019}. In addition to evolutionary strategies, reinforcement learning has been shown to be a viable solution for concurrent design of 2D simulated locomotive systems \cite{Ha2019j}. This is further shown to be a viable method by demonstrating more complex morphology modifications in 3D reaching and locomotive tasks \cite{Schaff2019e}. However, these techniques have not yet been applied to flexible systems for locomotive tasks. 

\section{Pogo-stick Model}
\label{sec:pogo_model}

\begin{figure}[t]
        \begin{center}
        \includegraphics[width=2.5cm]{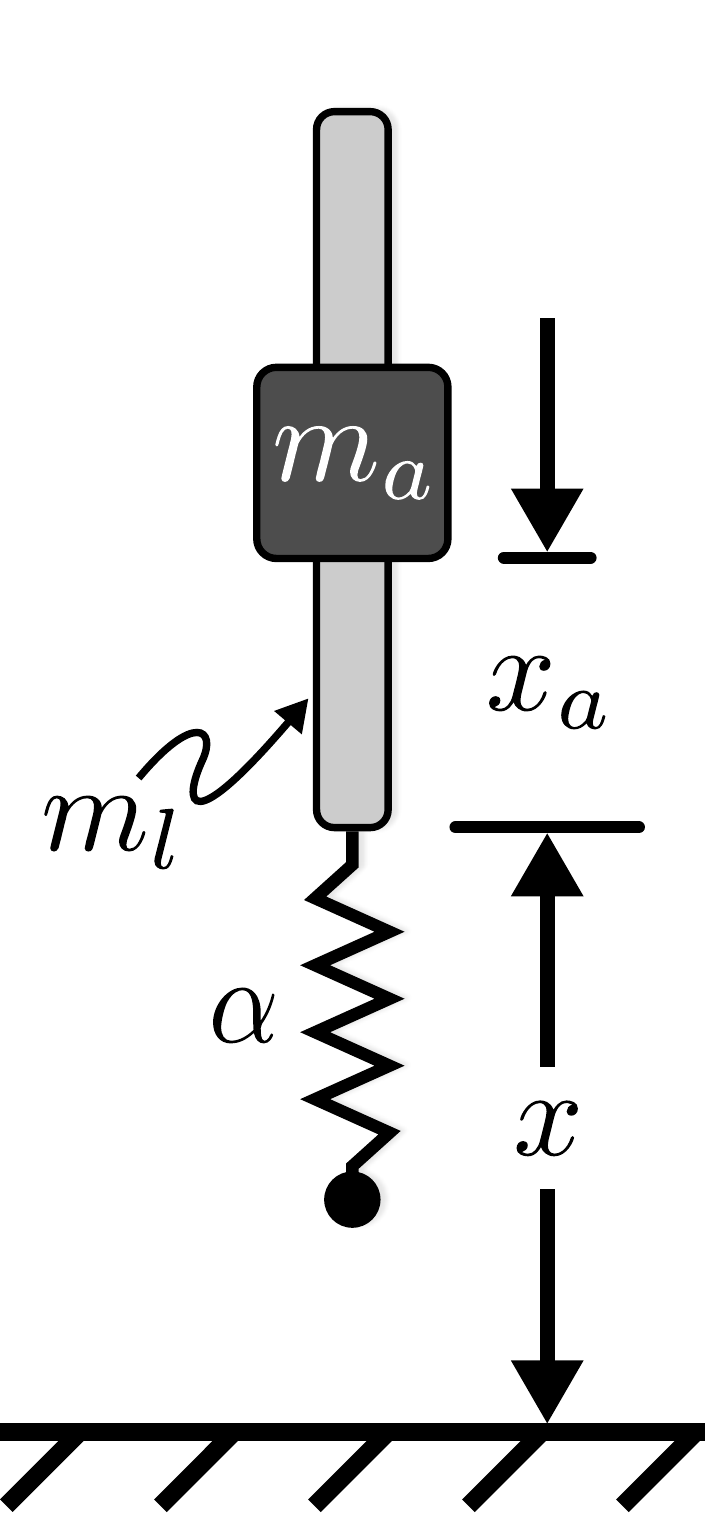}
        \caption{Pogo-stick System}
        \label{fig:pogoStickSystem} 
        \end{center}
        % \vspace{-4mm}
        \end{figure}

The pogo-stick model show in Figure~\ref{fig:pogoStickSystem} has been shown to be useful as a representation of several different running and jumping gaits \cite{Blickhan1993a}. As such, it is used in this work to demonstrate the ability of reinforcement learning for the mechanical design steps of concurrent design. The model parameters used in the simulations in this paper are summarized in Table~\ref{tab:pogoStickSystem}.
The variable $m_a$ represents the mass of the actuator, which moves along the rod with mass $m_l$. A nonlinear spring shown in the figure by constant $\alpha$ is used to represent flexibility within the jumping system. A damper (not shown in Figure~\ref{fig:pogoStickSystem}), is parallel to the spring. Variables $x$ and $x_a$ represent the system's vertical position with respect to the ground and the actuator's position along the rod, respectively. The system is additionally constrained such that it only moves vertically, therefore a controller is not required to balance the system.
	
The equations of motion describing the system are:
\begin{equation}
		\ddot{x} = \frac{\gamma}{m_t} \left(\alpha\,x + \beta\,x^3 + c\,\dot{x}\right)-\frac{m_a}{m_t}\,\ddot{x}_a-g
\end{equation}
where $x$ and $\dot{x}$ are position and velocity of the rod, respectively, the acceleration of the actuator, $\ddot{x}_a$, is the control input, and $m_t$ is the mass of the complete system. Constants $\alpha$ and $c$ represent the nonlinear spring and damping coefficient, respectively, and constant $\beta$ is set to $1e8$. Ground contact determines the value of $\gamma$, so that the spring and damper do not supply force while the leg is airborne:
	\begin{equation}
		\gamma =
		\left\{\begin{matrix}
		   -1, & x \leq 0\\ 
		   \hphantom{-} 0, & \mbox{otherwise}
		   \end{matrix}\right.
                %    \vspace{.25cm}
	 \end{equation}

Additionally, a spring compression limit was defined so that the spring could only compress to a specified amount. For this work, the spring was allowed to compress 0.008~m.
\begin{table}[t]
        \caption{Pogo-stick Model Parameters}
        \vspace{-4mm}
        \label{tab:pogoStickSystem}
        \begin{center}
                \begin{tabular}{|c||c|}
                %			& & \\ % put some space after the caption
                \hline
                \textbf{Model Parameter} & \textbf{Value}\\
                \hline
                Mass of Leg, $m_l$                                       & 0.175 kg                                          \\
                Mass of Actuator, $m_a$                                  & 1.003 kg                                          \\
                Spring Constant, $\alpha_{nominal}$                      & 5760 $\textup{N}/\textup{m}$                      \\
                Natural Frequency, $\omega_n$                            & $\sqrt{\frac{\alpha}{m_l + m_a}}$                 \\
                Damping Ratio, $\zeta_{nominal}$                         & 1e-2 \& 7.5e-2 $\frac{\textup{N}}{\textup{m/s}}$  \\
                % Damping Constant, $c$                                    & $2 \, \zeta \, \omega_n m$                        \\
                \hline
                Actuator Stroke, $(x_{a})_{\textup{max}}$                & 0.008 $\textup{m}$                                \\
                Max.\ Actuator Velocity, $(\dot{x}_{a})_{\textup{max}}$  & 1.0 $\textup{m}/\textup{s}$                       \\ 
                Max.\ Actuator Accel., $(\ddot{x}_{a})_{\textup{max}}$   & 10.0 $\textup{m}/\textup{s}^2$                    \\
                \hline
                \end{tabular}
        \end{center}
        % \vspace{-5mm}
\end{table}
% 
        
%%%%%%%%%%%%%%%%%%%%%%%%%%%%%%%%%%%%%%%%%%%%%%%%%%%%%%%%%%%%%%%%%%%%%%%%
\section{Jumping Command Design}
\label{sec:control_input_exp}

\begin{figure}[tb]
\begin{center}
\includegraphics[width = 3in]{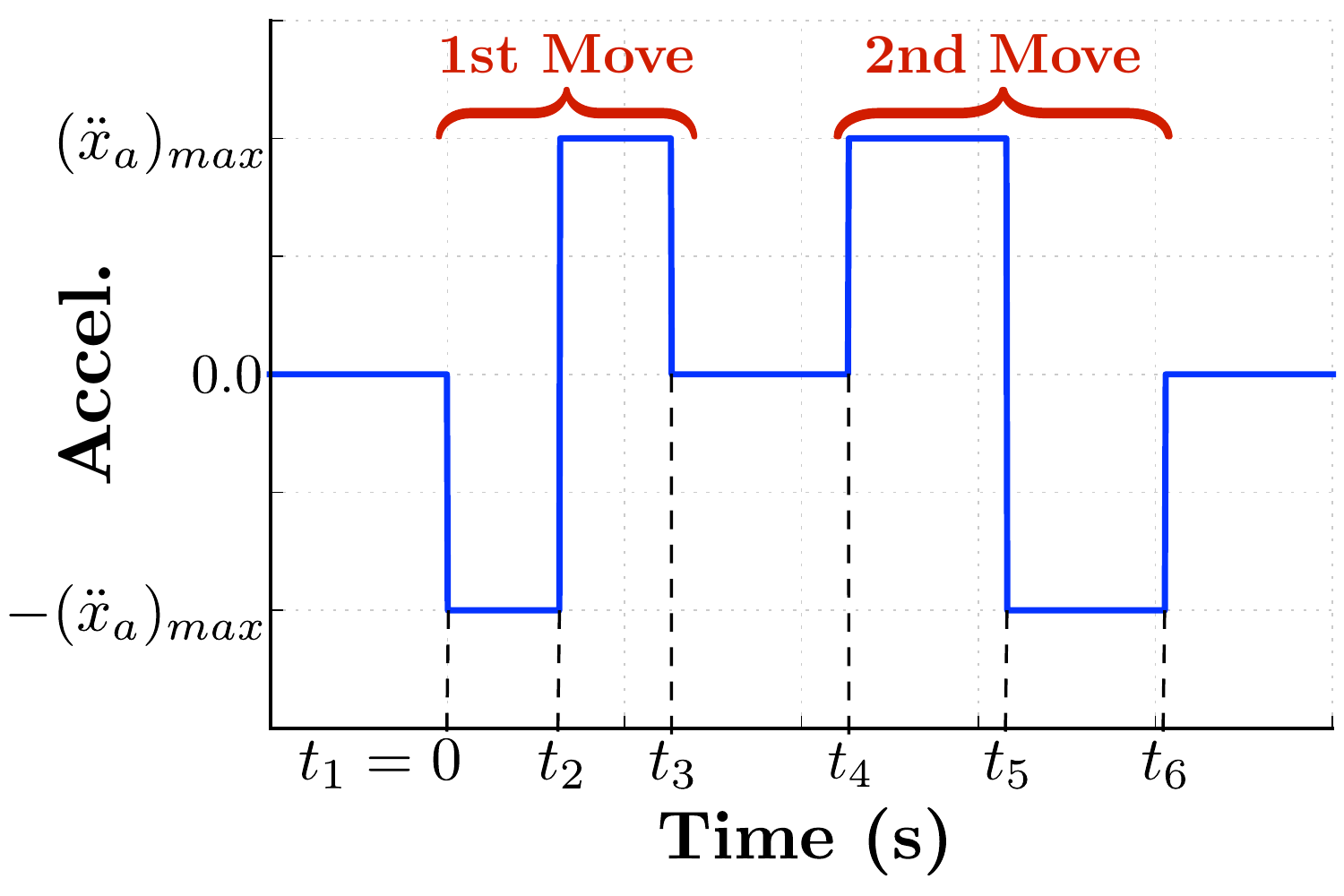}  
\caption{Jumping Command}
\label{fig:sim_command}
\end{center}
% \vspace{-4mm}
\end{figure}

\begin{figure}[tb]
\begin{center}
\includegraphics[width = 3in]{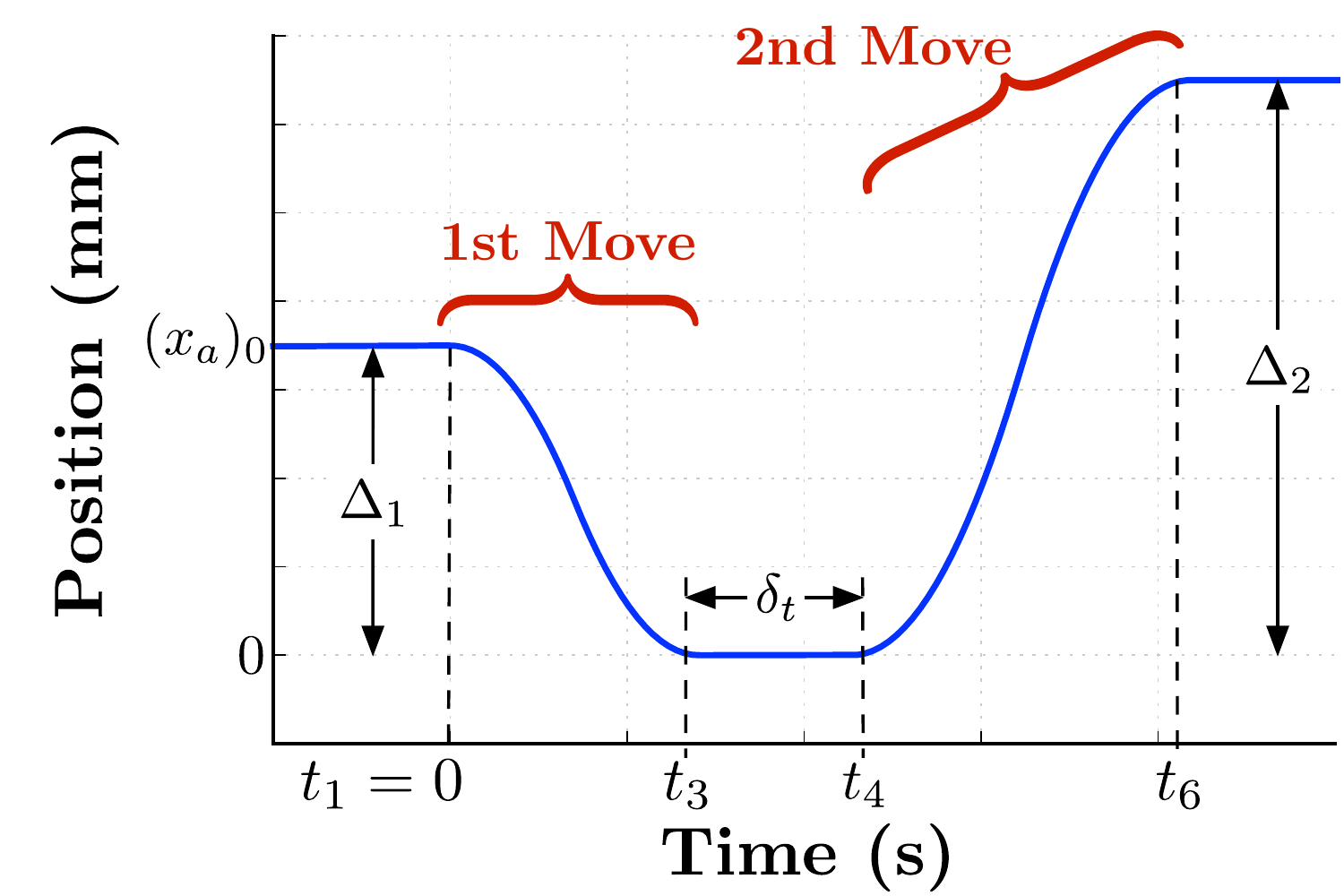}  
\caption{Resulting Actuator Motion}
\label{fig:command_act_motion}
\end{center}
% \vspace{-4mm}
\end{figure}

Bang-bang based jumping commands like the one shown in Figure \ref{fig:sim_command} are likely to result in a maximized jump height \cite{Vaughan2013}. For this command, the actuator mass travels at maximum acceleration within its allowable range, pauses, then accelerates in the opposite direction. Commands designed to complete this motion are bang-bang in each direction, with a selectable delay between them. The resulting motion of the actuator along the rod is shown in Figure \ref{fig:command_act_motion}. Starting from an initial position, $(x_a)_0$, it moves through a motion of stroke length $\Delta_1$, pauses there for $\delta_t$, then moves a distance $\Delta_2$ during the second portion of the acceleration input.


\begin{figure}[tbp]
\begin{center}
\includegraphics[width = 3.0in]{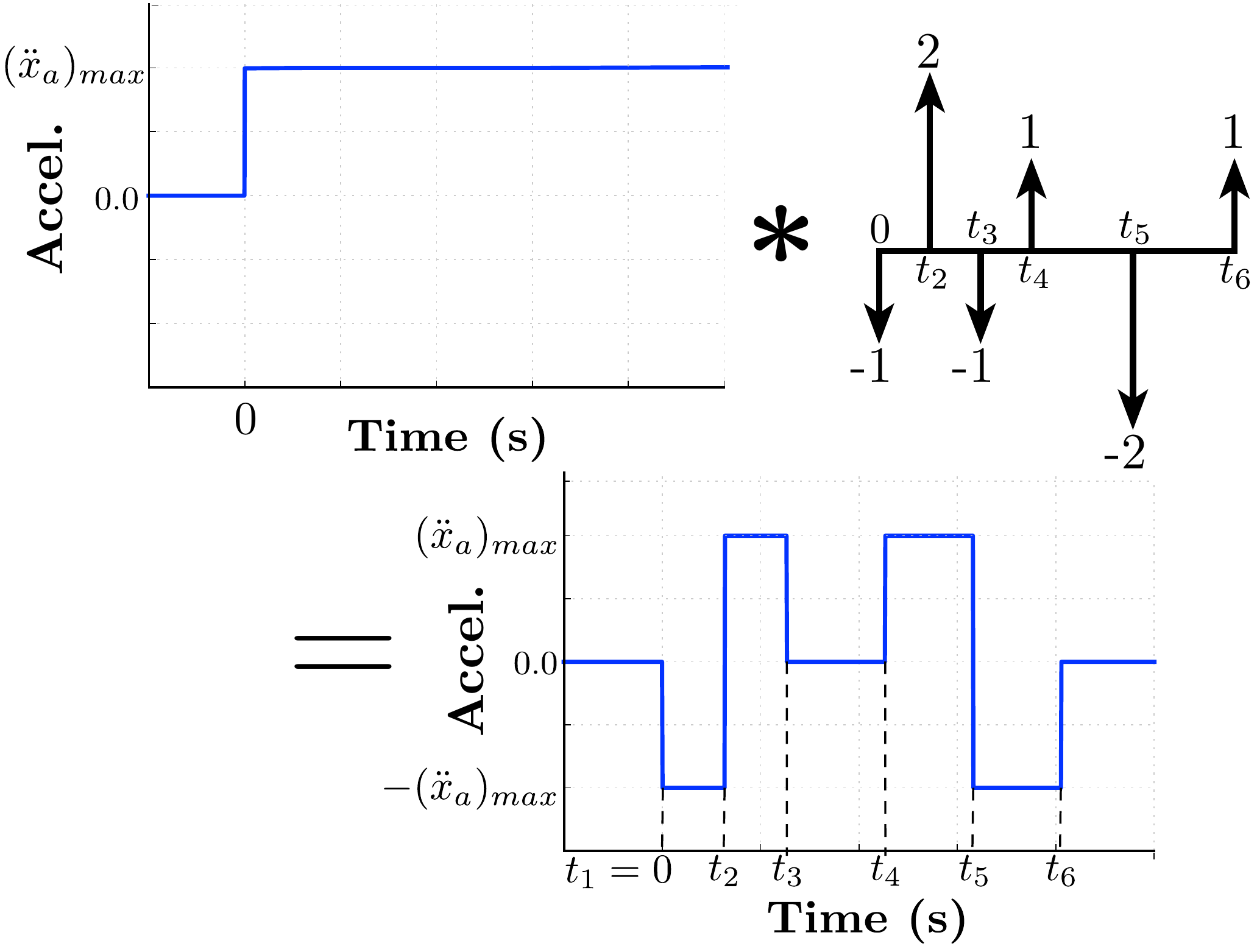}
\caption{Decomposition of the Jump Command into a Step Convolved with an Impulse Sequence}
\label{fig:jump_convolve}
\end{center}
% \vspace{-4mm}
\end{figure}

\begin{figure}[t]
        \begin{center}
        \includegraphics[width=3in]{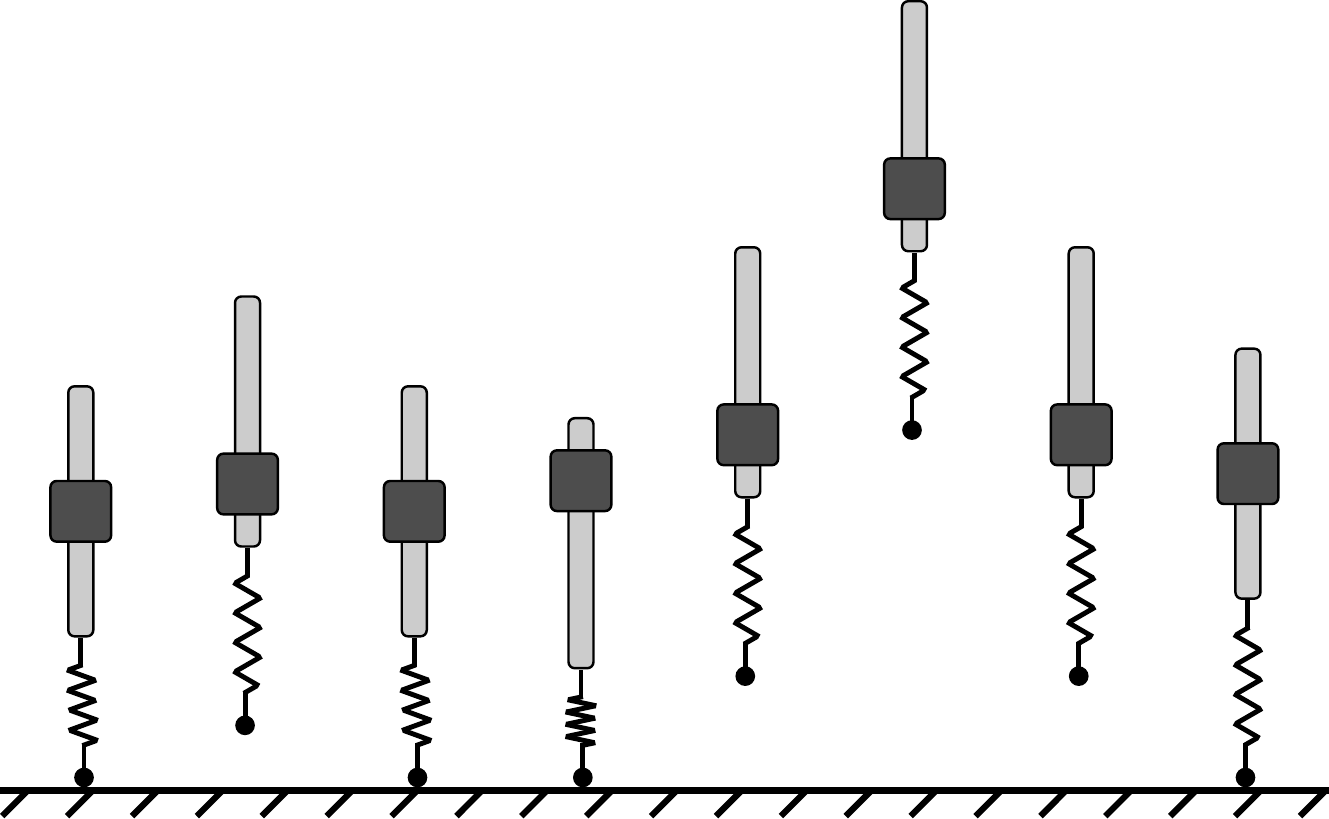}
        \caption{Example Stutter Jump}
        \label{fig:stutterJumpFigure} 
        \end{center}
        % \vspace{-4mm}
\end{figure}

This bang-bang-based profile can be represented as a step command convolved with a series of impulses, as shown in Figure \ref{fig:jump_convolve} \cite{Sorensen2008CommandinducedVA}. Using this decomposition, input-shaping principles and tools can be used to design the impulse sequence \cite{Singer:90, Singhose:94a}. 
For the bang-bang-based jumping command, the amplitudes of the resulting impulse sequence are fixed, $A_i = [-1, 2, -1, 1, -2, 1]$. The impulse times, $t_i$, can be varied and optimal selection of them can lead to a maximized jump height of the pogo-stick system \cite{Vaughan2013}. Commands of this form will often result in a stutter jump like what is shown in Figure~\ref{fig:stutterJumpFigure}, where the small initial jump allows the system to compress the spring to store energy to be used in the final jump. This jumping command type was used as the input for the pogo-stick during the simulation phase of training.

%%%%%%%%%%%%%%%%%%%%%%%%%%%%%%%%%%%%%%%%%%%%%%%%%%%%%%%%%%%%%%%%%%%%%%%%
\section{Learning Spring Constant and Damping Ratio}
\label{sec:learning_mech_params}
\subsection{Reinforcement Learning Algorithm}
The algorithm used for this work was Twin Delayed Deep Deterministic Policy Gradient (TD3) \cite{Fujimoto2018d}. This is an actor-critic algorithm wherein there exists two main neural networks and a set of twin trailing networks. The first main network is the actor, which determines the action of the agent. This network takes in the systems state, $\mathcal{S}$, and outputs the action, $\mathcal{A}$, based on the state. The critic is an estimator of the value of being in a state and is used to determine the difference between expected and estimated value used to update the actor network during training. It takes in the systems state, $\mathcal{S}$, and outputs the expected future reward, $\mathbb{R}$, from being in that state. The twin trailing networks are used to find the temporal difference error against the critic network which is used to update the critic network.

The training hyperparameters were selected based on TD3's author recommendations and Stable Baselines experimental findings and are displayed in Table~\ref{tab:training_hyperameters}. All of the hyperparameters, with the exception of the rollout (Learning Starts) and the replay buffer, were set according to Stable Baselines standards. The rollout setting was defined such that the agent could search the design space at random, filling the replay buffer with enough experience to prevent the agent from converging to a design space that was not optimal. The replay buffer was sized proportional to the number of training steps due to system memory constraints.  
\begin{table}[t]
        \caption{TD3 Training Hyperparameters}
        \vspace{-4mm}
        \label{tab:training_hyperameters}
        \begin{center}
        \begin{tabular}{|c||c|}
        \hline
        Hyperameter                     & Value                           \\
        \hline
        Learning Rate, $\alpha$         & 0.001                           \\
        Learning Starts                 & 100 Steps                       \\
        Batch Size                      & 100 Transitions                 \\
        Tau, $\tau$                     & 0.005                           \\
        Gamma, $\gamma$                 & 0.99                            \\
        Training Frequency              & 1:Episode                       \\
        Gradient Steps                  & $\propto$ Training Frequency    \\
        Action Noise,  $\epsilon$       & None                            \\
        Policy Delay                    & 1 : 2 Q-Function Updates        \\
        Target Policy Noise, $\epsilon$ & 0.2                             \\
        Target Policy Clip, $c$         & 0.5                             \\
        Seed                            & 100 Random Seeds                \\
        \hline
        \end{tabular}
        \end{center}
        \vspace{-5mm}
\end{table}
\subsection{Training Environment Design}
To allow the agent to find a mechanical design, a reinforcement learning environment conforming to the OpenAI Gym standard \cite{Brockman2016c} was created for the pogo-stick model described in Section~\ref{sec:pogo_model}, including a fixed controller input based on the algorithm described in the previous section. Unlike the common use case for RL, which is tasking the agent with finding a control input to match a design, the agent in this work was tasked with finding mechanical parameters to match a control input.

The mechanical parameters the agent was tasked with optimizing were the spring constant and the damping ratio of the pogo-stick system. At each episode during training, the agent selected a set of design parameters from a distribution of available designs. The actions applied, $\mathcal{A}$, and transitions saved, $\mathcal{S}$, from the environment were defined as follows:
\begin{equation}
        \label{eq:action}
        \begin{aligned}
        \mathcal{A} = \{ \{ a_{\alpha} \in \mathbb{R}: [-0.9 \alpha, 0.9 \alpha] \}, \\ 
        \{ a_{\zeta} \in \mathbb{R}: [-0.9 \zeta, 0.9 \zeta] \} \}
        \end{aligned}
\end{equation} 
\begin{equation}
        \label{eq:transitions}
        \mathcal{S}= \left \{\sum_{t=0}^{t_f}x_t,~ \sum_{t=0}^{t_f}\dot{x}_t,~ \sum_{t=0}^{t_f}x_{at},~ \sum_{t=0}^{t_f}\dot{x}_{at} \right \}
\end{equation}
where $\alpha$ and $\zeta$ are the nominal spring constant and damping ratio of the pogo-stick, respectively; $x_t$ and $\dot{x}_t$ are the pogo-stick rod height and velocity steps, and $x_{at}$ and $\dot{x}_{at}$ are the pogo-stick actuator position and velocity steps, all captured during simulation. 

\subsection{Reward Function Design}

The RL algorithm was utilized to find designs for two different reward cases. Time series data was captured during the simulation phase of training and was used to evaluate the designs performance through these rewards. The first reward case used was:
\begin{equation}
        \mathbb{R}_1 = \left (\sum_{t=0}^{t_f}x_t  \right )_{max}
\end{equation}
where $x_t$ was the pogo-stick's rod height at each step during simulation. The goal of the first reward was to find a design that would cause the pogo-stick to jump as high as possible. 

The reward for the second case was:
\begin{equation}
        \mathbb{R}_2 = \frac{1}{\frac{\left | \mathbb{R}_1 - x_s \right |}{x_s} + 1}
\end{equation}
where $x_s$ was the desired jump height, which was set to 0.01~m. The second case was utilized to test RL's ability to find a design that minimized the error between the maximum height reached and the desired maximum height to reach. 

\subsection{Training Schedule}

To evaluate the algorithm's ability to robustly find design parameters meeting performance needs regardless of the neural network initializations, 100 different agents were trained with different network initialization seeds. The RL agents were trained in two different environments, one with a narrow range of allowable damping ratios and one with a wider range of possible damping ratios.

The number of episodes that were performed was 1000, with the first 100 being rollout steps. This provided the agents, in both cases previously mentioned, enough learning time to converge to designs that satisfied the performance requirements. During the training process, the height reached during the simulation phase (per environment step) and the design parameters selected by the algorithm where collected to evaluate the learning process. 

%%%%%%%%%%%%%%%%%%%%%%%%%%%%%%%%%%%%%%%%%%%%%%%%%%%%%%%%%%%%%%%%%%%%%%%%
\section{Jumping Height Reached} 
\label{sec:results}

\subsection{Design Space Performance}
\begin{figure}[!t]
        \begin{center}
        \includegraphics[width = 3in]{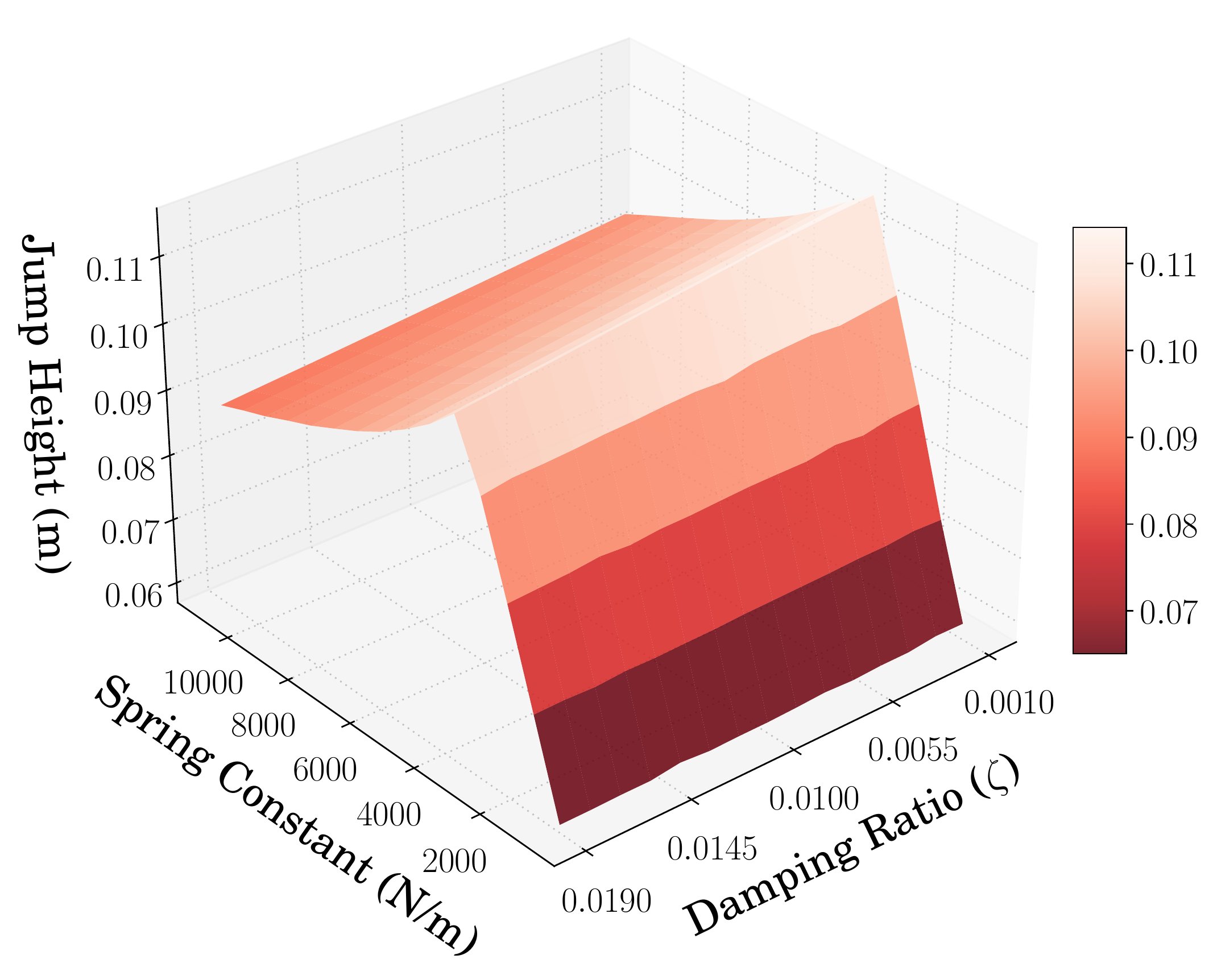}  
        \caption{Jumping Performance of Narrow Design Space}
        \label{fig:spring_zeta_height_close}
        \end{center}
        % \vspace{-5mm}
\end{figure}
\begin{figure}[!t]
        \begin{center}
        \includegraphics[width = 3in]{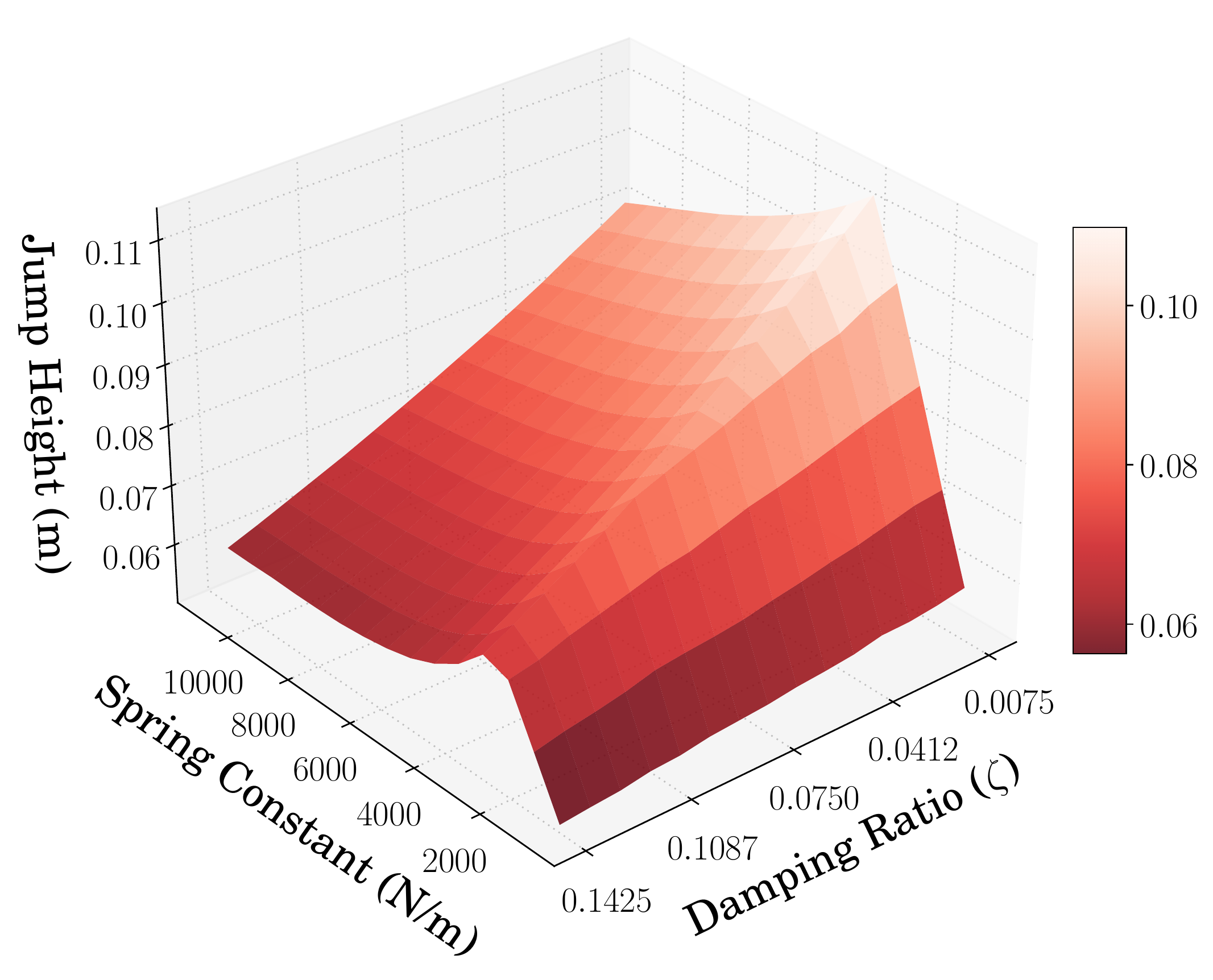}  
        \caption{Jumping Performance of Broad Design Space}
        \label{fig:spring_zeta_height_far}
        \end{center}
        % \vspace{-5mm}
\end{figure}

Figures~\ref{fig:spring_zeta_height_close} and~\ref{fig:spring_zeta_height_far} represent the heights the pogo-stick could reach for the two different design spaces. The design space provided for the first case, shown in Figure~\ref{fig:spring_zeta_height_close}, represents a space where the allowable damping ratio was limited to a fairly narrow range. This limits the solution space, making it less likely that the agent will settle to a locally optimal value. The design space provided for the second case, shown in Figure~\ref{fig:spring_zeta_height_far}, represents a space where a wider range of damping ratios are allowed. This wider range of possible values makes it more likely that the agent will settle to a local maxima.

\subsection{Design Learned Given Narrow Design Space}

\begin{figure}[tb]
        \begin{center}
        \includegraphics[width = 3in]{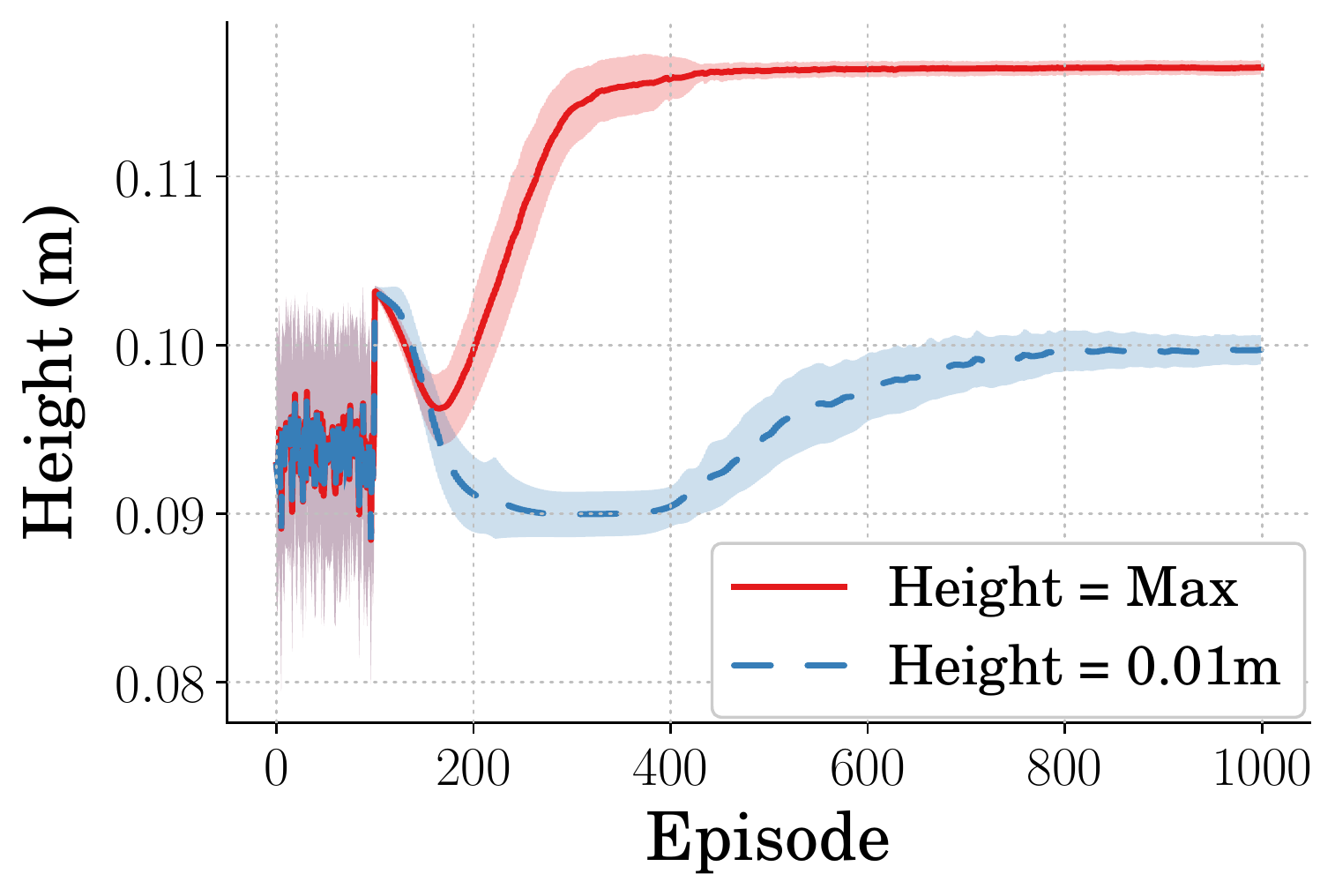}  
        \caption{Height Reached During Training}
        \label{fig:height_vs_step_close}
        \end{center}
        % \vspace{-4mm}
        \end{figure}

\begin{figure}[tb]
        \begin{center}
        \includegraphics[width = 3in]{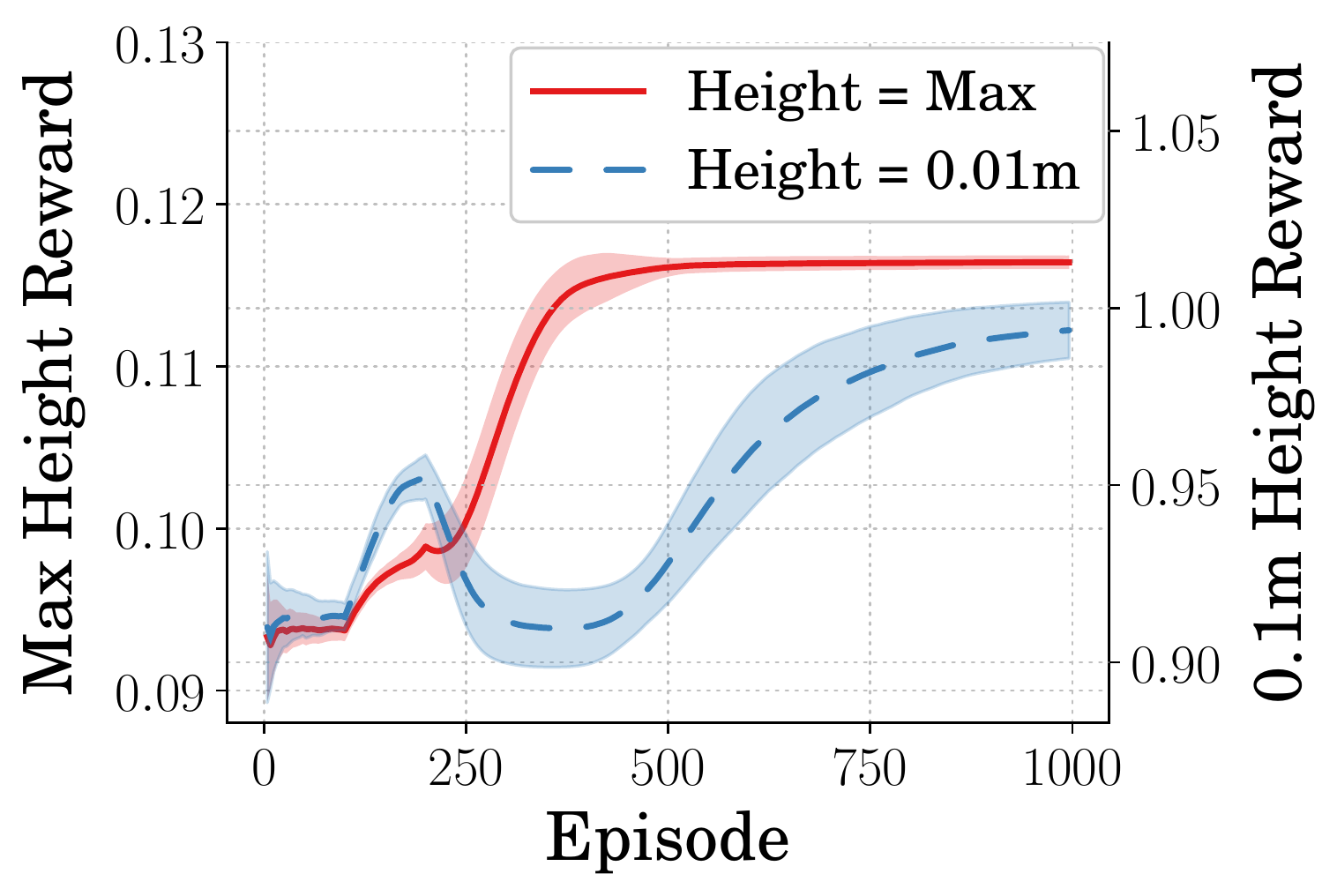}  
        \caption{Reward Received During Training}
        \label{fig:rew_vs_step_close}
        \end{center}
        % \vspace{-4mm}
        \end{figure}

Figure~\ref{fig:height_vs_step_close} shows the height achieved by the learned designs for the agents given the narrow range of possible damping ratio values. For the agents learning designs to maximize jump height, Figure~\ref{fig:height_vs_step_close} can be compared with Figure~\ref{fig:spring_zeta_height_close} showing that the agent learned a design nearing one which would achieve maximum performance. Additionally, looking at the agents learning designs to jump to the specified 0.01~m, the designs learned accomplish this with slightly more variance than that of the maximum height case. Figure~\ref{fig:rew_vs_step_close} shows the rewards the agents received during training and support that both agent types learned designs which converged. 

\begin{figure}[tb]
        \begin{center}
        \includegraphics[width = 3in]{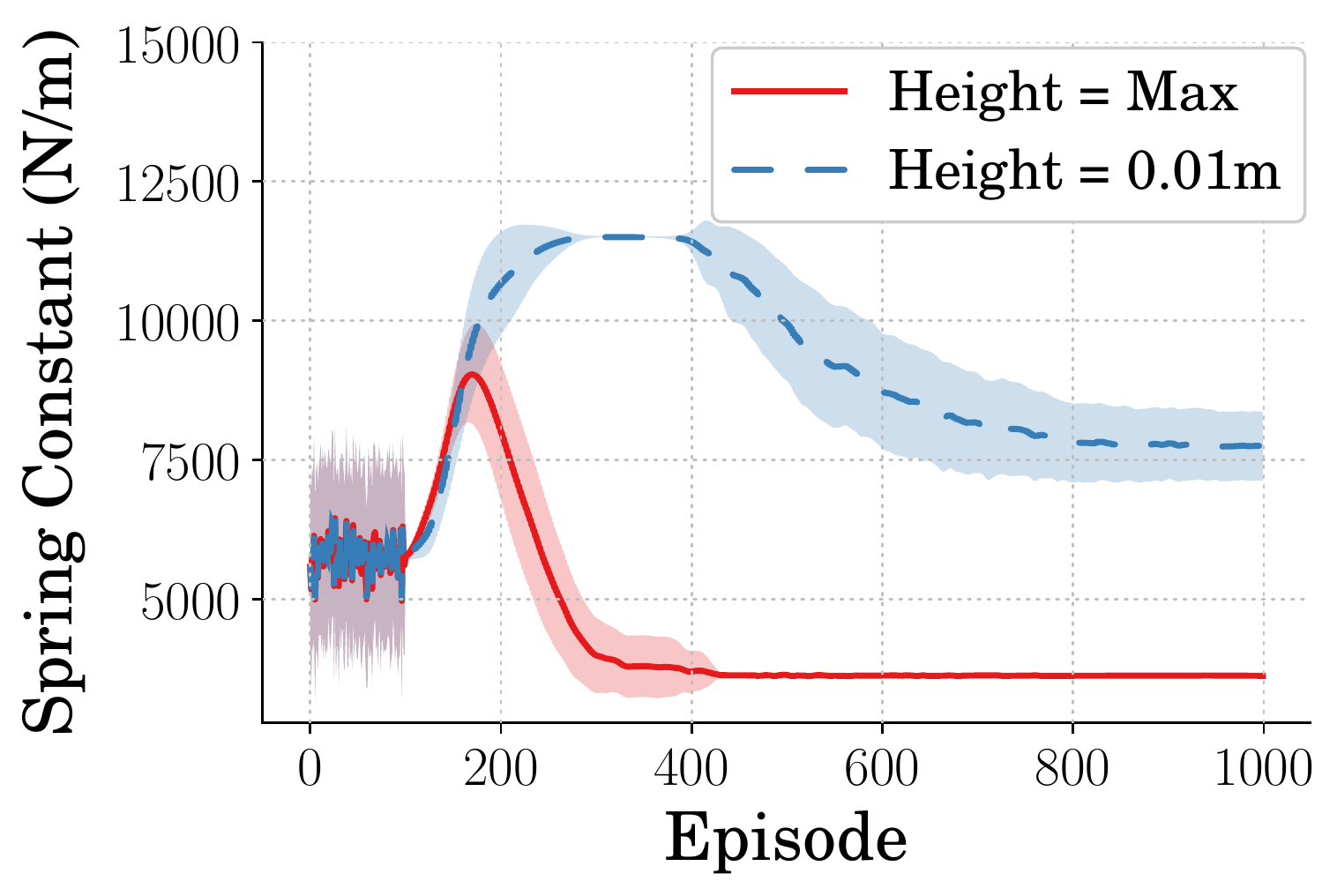}  
        \caption{Spring Constant Selected During Training}
        \label{fig:spring_vs_step_close}
        \end{center}
        % \vspace{-4mm}
        \end{figure}

\begin{figure}[tb]
        \begin{center}
        \includegraphics[width = 3in]{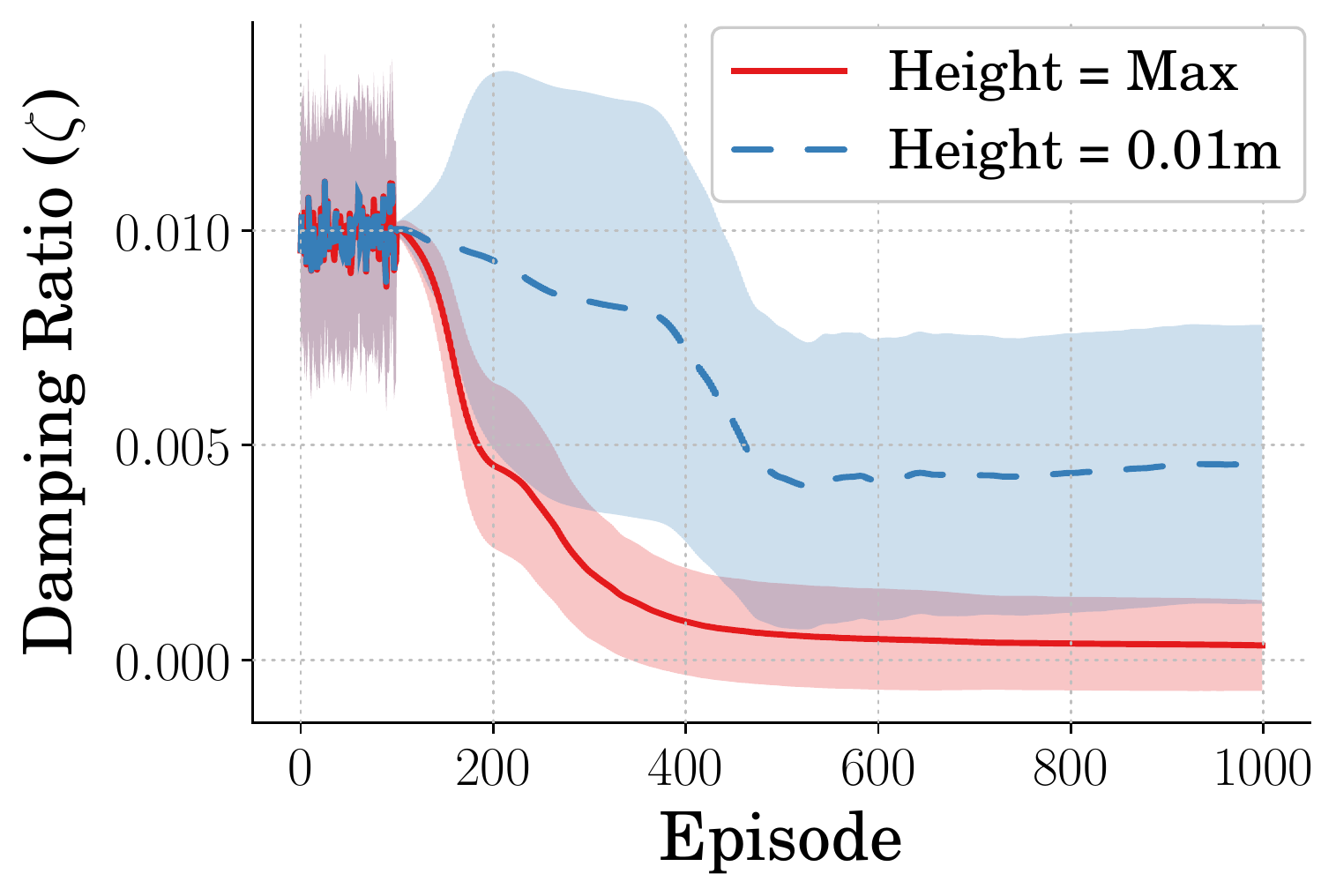}  
        \caption{Damping Ratio Selected During Training}
        \label{fig:zeta_vs_step_close}
        \end{center}
        % \vspace{-4mm}
        \end{figure}

The average and standard deviation of the spring constant and damping ratio design parameters the agents selected during training are shown in Figures~\ref{fig:spring_vs_step_close} and~\ref{fig:zeta_vs_step_close}. These plots represent the learning curves for the agents learning design parameters to maximize jump height and the agents learning design parameters to jump to 0.01~m. There is a high variance in both the spring constant and the damping ratio found for the agents that learned designs to jump to a specified height. The agents which were learning designs which maximized height found designs with very little variance in terms of spring constant and significantly less variances in terms of damping ratio.

\subsection{Design Learned Given Broad Design Space}

\begin{figure}[t]
        \begin{center}
        \includegraphics[width = 3in]{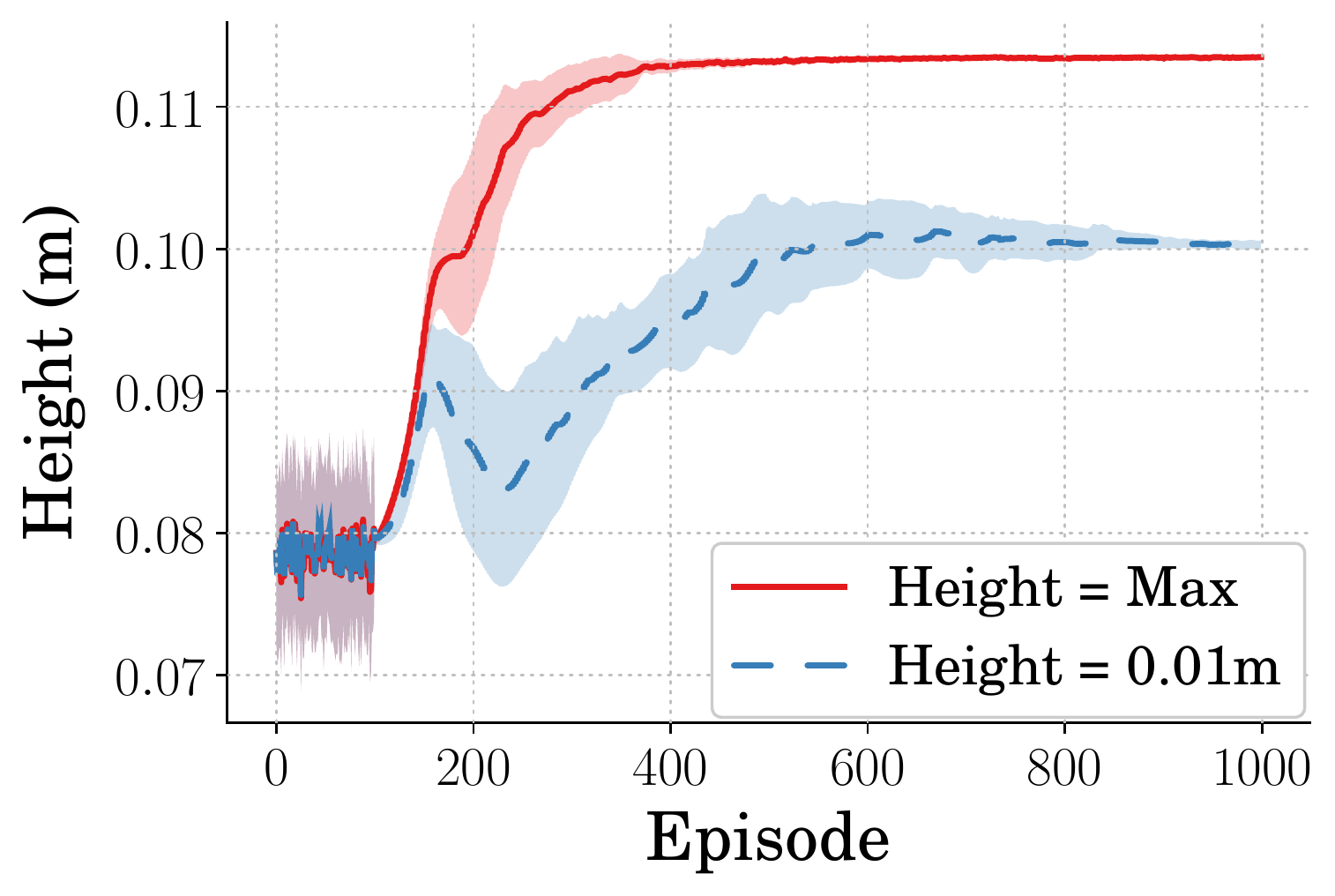}  
        \caption{Height Reached During Training}
        \label{fig:height_vs_step_far}
        \end{center}
        % \vspace{-4mm}
        \end{figure}

\begin{figure}[t]
        \begin{center}
        \includegraphics[width = 3in]{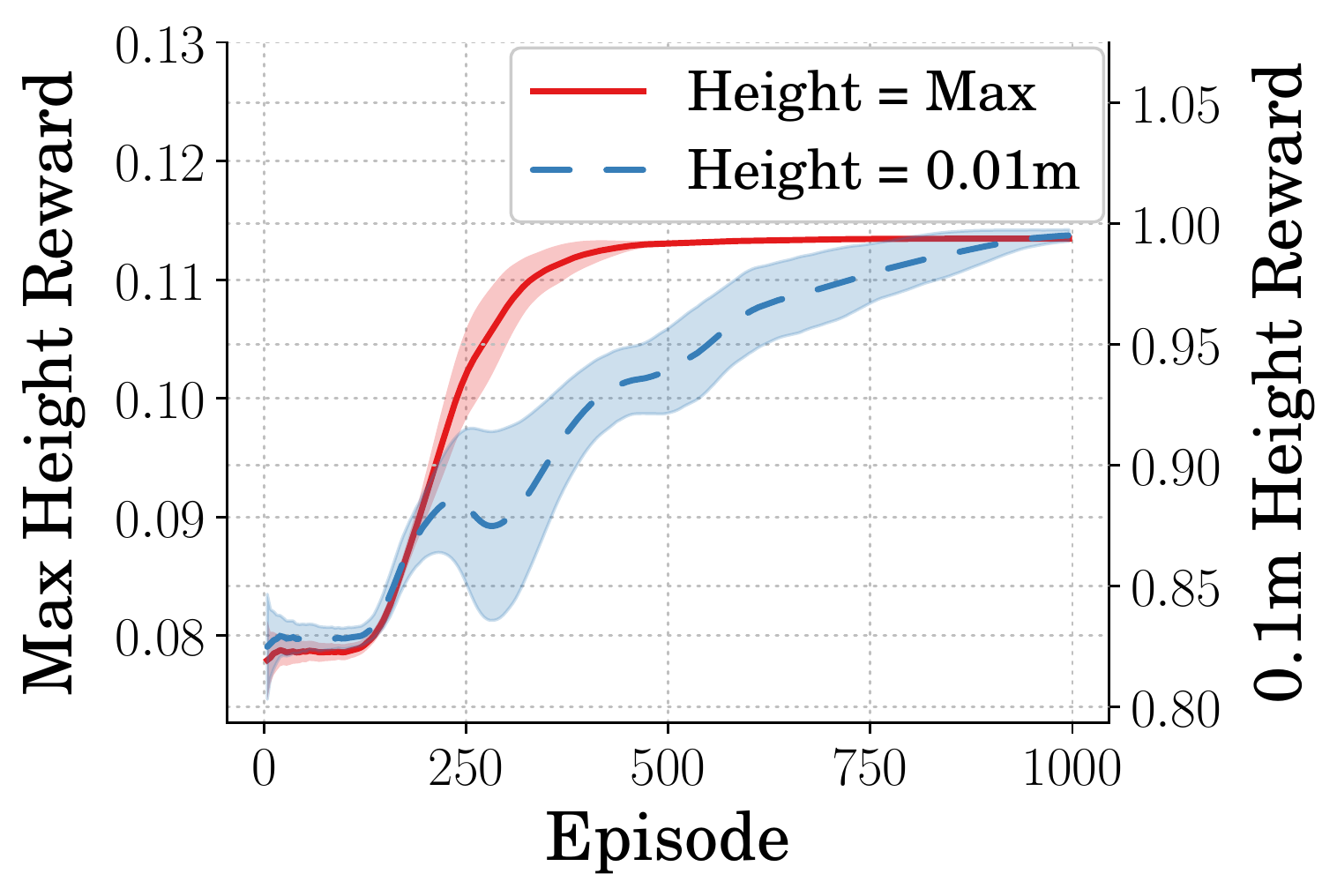}  
        \caption{Reward Received During Training}
        \label{fig:rew_vs_step_far}
        \end{center}
        % \vspace{-4mm}
        \end{figure}

Figure~\ref{fig:height_vs_step_far} shows the height achieved by the learned designs for the agents given a wider range of damping ratios. For the agents learning designs to maximize jump height, Figure~\ref{fig:height_vs_step_far} can be compared with Figure~\ref{fig:spring_zeta_height_far} showing that the agents learned a design nearing one which would achieve maximum performance. Additionally, looking at the agents learning designs to jump to the specified 0.01~m, the designs learned accomplish this, only with slightly more variance than what is seen in the maximum height agents. Figure~\ref{fig:rew_vs_step_far} shows the rewards the agents received during training and support that both agent types learned designs which converged. In this case though, the agent learning a design to jump to a specific height requires more learning steps to converge. 

\begin{figure}[tb]
        \begin{center}
        \includegraphics[width = 3in]{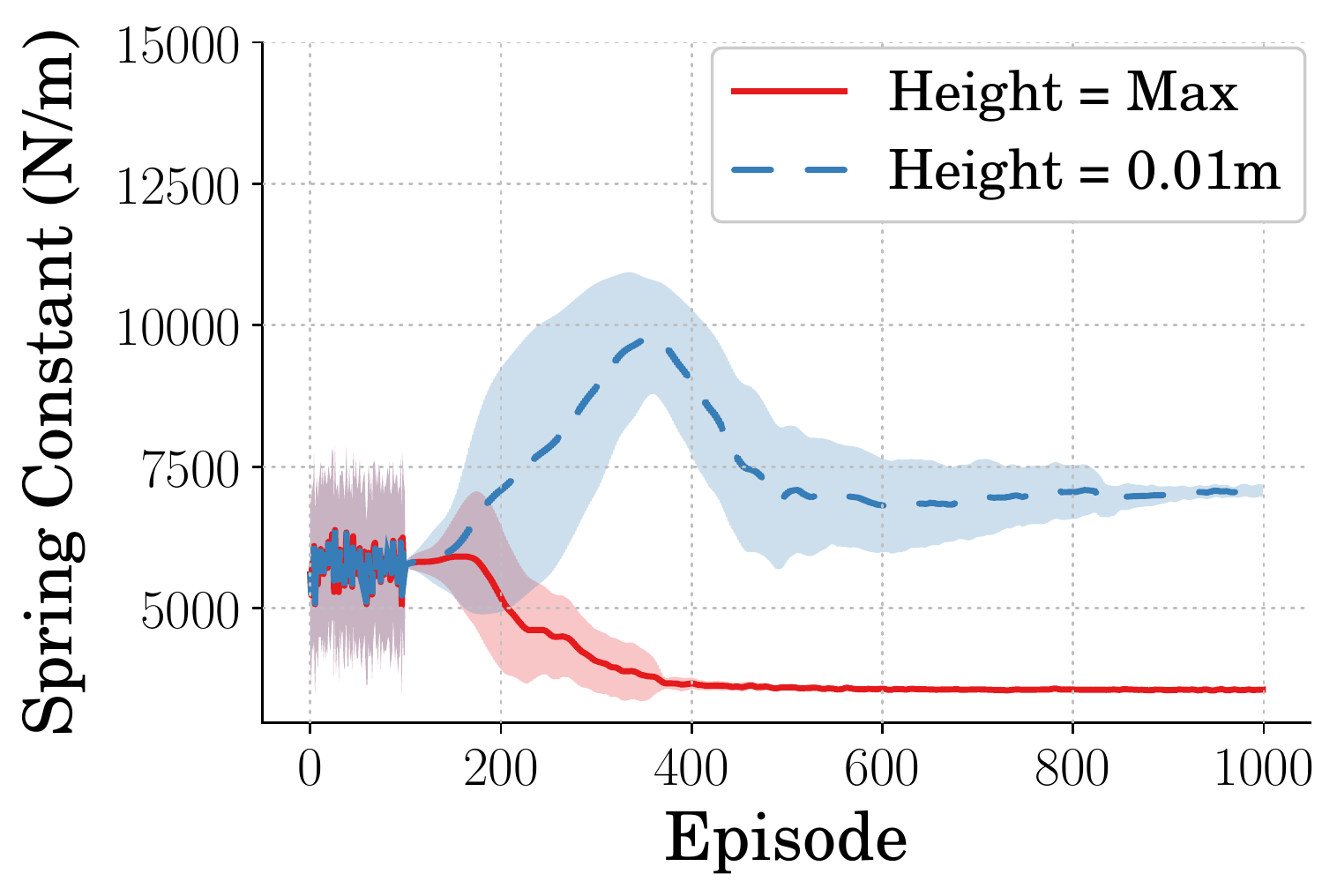}  
        \caption{Spring Constant Selected During Training}
        \label{fig:spring_vs_step_far}
        \end{center}
        % \vspace{-4mm}
        \end{figure}

\begin{figure}[tb]
        \begin{center}
        \includegraphics[width = 3in]{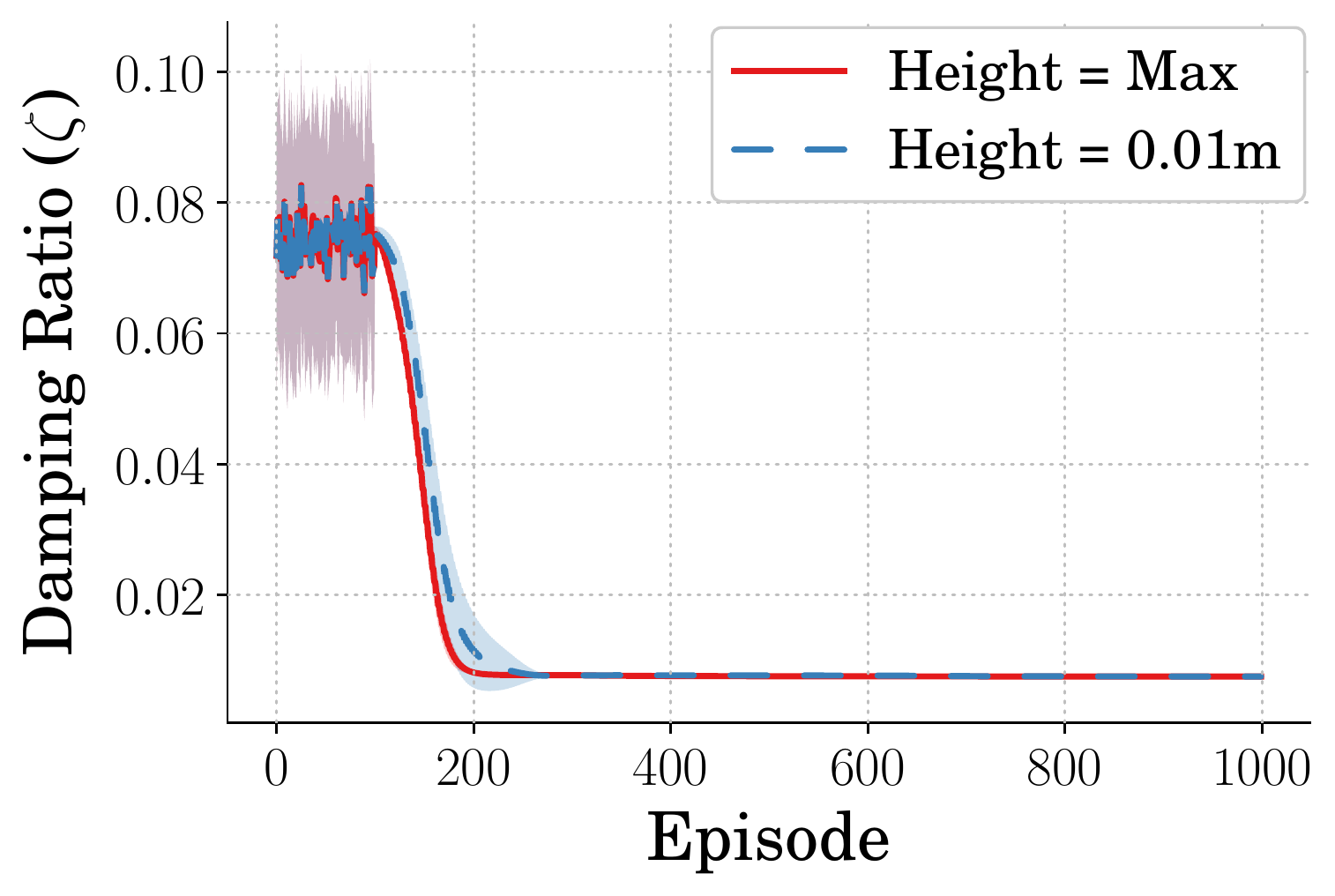}  
        \caption{Damping Ratio Selected During Training}
        \label{fig:zeta_vs_step_far}
        \end{center}
        % \vspace{-4mm}
        \end{figure}

The average and standard deviation of the spring constant and damping ratio design parameters the agents selected during training are shown in Figures~\ref{fig:spring_vs_step_far} and~\ref{fig:zeta_vs_step_far}. For the agents that learned designs to jump to a specified height, it can be seen that there is a high variance in spring constant throughout training. However, the majority of agents converge to a specific design, lowering the variance. The same can be seen in the damping ratio; however, the variance is mitigated significantly earlier in training. The agents which were learning designs that maximized height found them with very little variance in terms of spring constant and damping ratio. 


\subsection{Average Design Performance}
\begin{figure}[tb]
        \begin{center}
        \includegraphics[width = 3in]{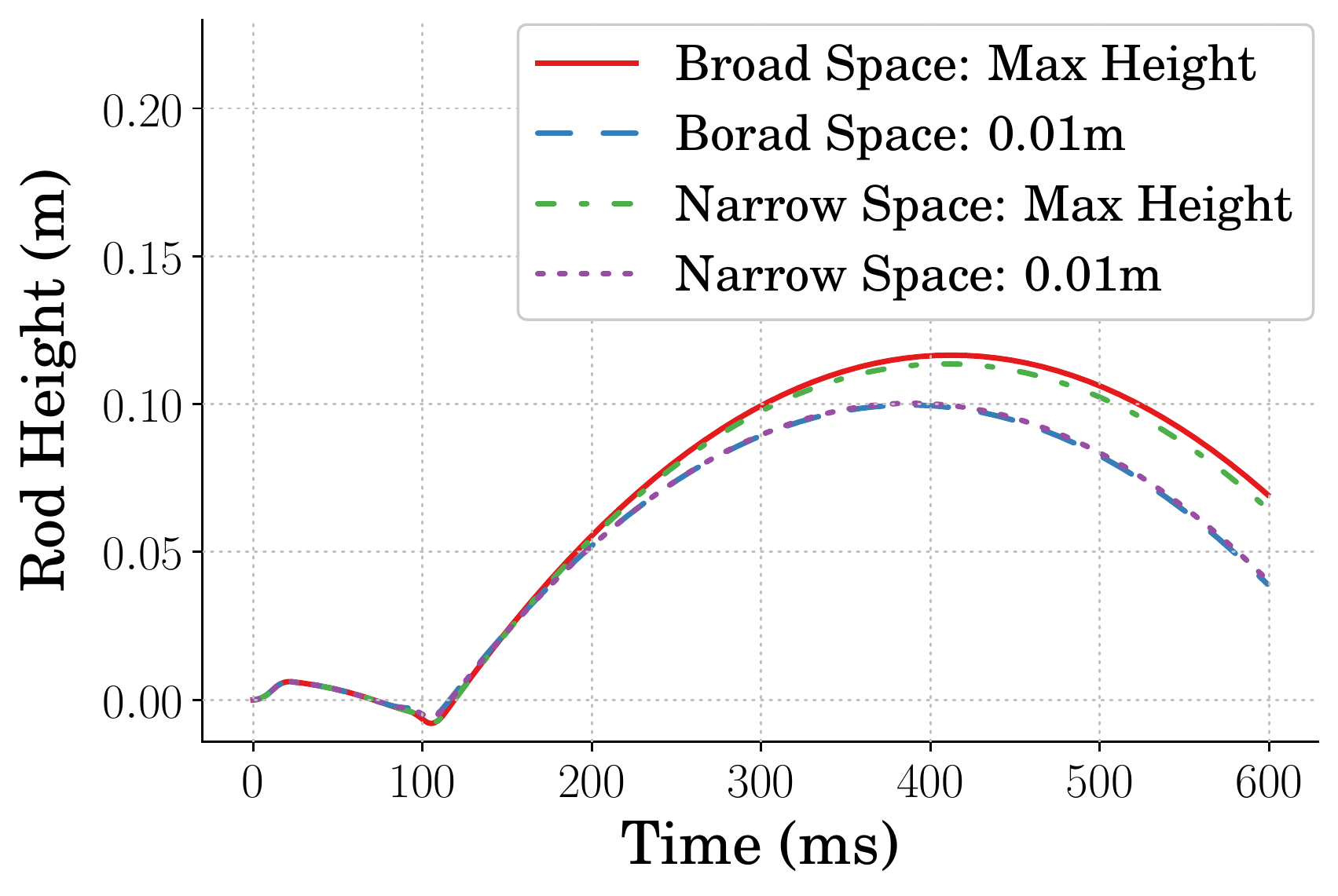}  
        \caption{Height vs Time of Average Optimal Designs}
        \label{fig:height_vs_time}
        \end{center}
        % \vspace{-4mm}
        \end{figure}

The final mean and standard deviation of the design parameters for the two different cases are presented in Table~\ref{tab:learned_design_params}. Figure~\ref{fig:height_vs_time} shows the jumping performance of the mean designs learned for both cases tested. The agents tasked with finding designs to jump to the specified 0.01~m, did so with minimal error. The difference seen in maximum height reached between the two cases represents the difference in the damping ratio design space the agents had access to. The peak heights achieved can be compared again to Figures~\ref{fig:spring_zeta_height_close} and~\ref{fig:spring_zeta_height_far} to show that the agents learned designs nearing those achieving maximum performance.

\begin{table*}[ht]
        \caption{Learned Design Parameters}
        \vspace{-4mm}
        \label{tab:learned_design_params}
                \begin{center}
                \begin{tabular}{|c||c||c||c||c|}
                \hline
                \multicolumn{2}{|c||}{Training Case}                                                                           & Design Parameter & Mean     & STD      \\
                \hline
                \multirow{4}{*}{\centering Narrow Design Space}           & \multirow{2}{*}{\centering Learn Max Height}       & Spring Constant  & 3.62e03  & 3.82e01  \\
                                                                          &                                                    & Damping Ratio    & 3.37e-04 & 2.11e-03 \\
                                                                          & \multirow{2}{*}{\centering Learn Specified Height} & Spring Constant  & 7.74e03  & 1.24e03  \\
                                                                          &                                                    & Damping Ratio    & 4.55e-03 & 6.49e-03 \\
                \multirow{4}{*}{\centering Broad Design Space}            & \multirow{2}{*}{\centering Learn Max Height}       & Spring Constant  & 3.55e03  & 4.86e01  \\
                                                                          &                                                    & Damping Ratio    & 7.53e-03 & 8.86e-06 \\
                                                                          & \multirow{2}{*}{\centering Learn Specified Height} & Spring Constant  & 7.07e03  & 2.16e02  \\
                                                                          &                                                    & Damping Ratio    & 7.54e-03 & 3.27e-05 \\
                \hline
                \end{tabular}
                \end{center}
        \end{table*}

%%%%%%%%%%%%%%%%%%%%%%%%%%%%%%%%%%%%%%%%%%%%%%%%%%%%%%%%%%%%%%%%%%%%%%%%
\section{Conclusion}
\label{sec:conclusion}
The pogo-stick model was used in conjunction with a predetermined control input to determine if a reinforcement learning algorithm (TD3) could be used to find optimal performing design parameters regarding jumping performance. This work was done in part to determine if reinforcement learning could be used as the mechanical design learner for an intelligent concurrent design algorithm. It was shown that when providing an agent with a design space that was smaller in size, the agents performed well in finding design parameters which met the performance constraints. The designs found were high in design variance, however. It was additionally shown that when provided with larger design space, the agents excelled at finding design parameters which were lower in design variance but still met the design constraints. 

\addtolength{\textheight}{-7.5cm}   % This command serves to balance the column lengths

%%%%%%%%%%%%%%%%%%%%%%%%%%%%%%%%%%%%%%%%%%%%%%%%%%%%%%%%%%%%%%%%%%%%%%%%
\section*{ACKNOWLEDGMENT}

The authors would like to thank the Louisiana Crawfish Promotion and Research Board for their support of this work.

\bibliographystyle{IEEEtran} % We choose the &quot;plain&quot; reference style
\bibliography{ACC_LearnMechParams_2021} % Entries are in the &quot;refs.bib&quot; file</code></pre>

\end{document}